\documentclass[sigconf]{acmart}

\usepackage{soul}
\usepackage{hyperref}
\usepackage[utf8]{inputenc}
\usepackage{amsthm}
\usepackage{algorithmic}
\usepackage{multirow}

\usepackage{amsfonts,amssymb}
\usepackage{balance}
\usepackage[ruled,vlined,linesnumbered]{algorithm2e}
\AtBeginDocument{%
  \providecommand\BibTeX{{%
    \normalfont B\kern-0.5em{\scshape i\kern-0.25em b}\kern-0.8em\TeX}}}

\copyrightyear{2023}
\acmYear{2023}
\setcopyright{acmlicensed}
\acmConference[CIKM '23] {Proceedings of the 32nd ACM International Conference on Information and Knowledge Management}{October 21--25, 2023}{Birmingham, United Kingdom.}
\acmBooktitle{Proceedings of the 32nd ACM International Conference on Information and Knowledge Management (CIKM '23), October 21--25, 2023, Birmingham, United Kingdom}
\acmPrice{15.00}
\acmISBN{979-8-4007-0124-5/23/10}
\acmDOI{10.1145/3583780.3614962}

\begin{document}

\title{MemDA: Forecasting Urban Time Series with Memory-based Drift Adaptation}

\author{Zekun Cai}
\authornote{This work was fulfilled when Zekun Cai interned at Tencent.}
\email{caizekun@csis.u-tokyo.ac.jp}
\orcid{0000-0002-5773-1395}
\affiliation{%
  \institution{The University of Tokyo}
  \city{Tokyo}
  \country{Japan}
}

\author{Renhe Jiang}
\authornote{Corresponding Author}
\email{jiangrh@csis.u-tokyo.ac.jp}
\affiliation{%
  \institution{The University of Tokyo}
  \city{Tokyo}
  \country{Japan}
}

\author{Xinyu Yang}
\email{xinyuyang@tencent.com}
\affiliation{%
  \institution{Tencent Corporation}
  \city{Beijing}
  \country{China}
}

\author{Zhaonan Wang}
\email{znwang@csis.u-tokyo.ac.jp}
\affiliation{%
  \institution{The University of Tokyo}
  \city{Tokyo}
  \country{Japan}
}

\author{Diansheng Guo}
\authornotemark[2]
\email{diansheng.guo@gmail.com}
\affiliation{%
  \institution{Tencent Corporation}
  \city{Beijing}
  \country{China}
}

\author{Hill Hiroki Kobayashi}
\email{kobayashi@ds.itc.u-tokyo.ac.jp}
\affiliation{%
  \institution{The University of Tokyo}
  \city{Tokyo}
  \country{Japan}
}

\author{Xuan Song}
\email{songxuan@csis.u-tokyo.ac.jp}
\affiliation{%
  \institution{The University of Tokyo}
  \city{Tokyo}
  \country{Japan}
}

\author{Ryosuke Shibasaki}
\email{shiba@csis.u-tokyo.ac.jp}
\affiliation{%
  \institution{The University of Tokyo}
  \city{Tokyo}
  \country{Japan}
}

\renewcommand{\shortauthors}{Zekun Cai, et al.}

\begin{abstract}
    Urban time series data forecasting featuring significant contributions to sustainable development is widely studied as an essential task of the smart city. However, with the dramatic and rapid changes in the world environment, the assumption that data obey Independent Identically Distribution is undermined by the subsequent changes in data distribution, known as concept drift, leading to weak replicability and transferability of the model over unseen data. To address the issue, previous approaches typically retrain the model, forcing it to fit the most recent observed data. However, retraining is problematic in that it leads to model lag, consumption of resources, and model re-invalidation, causing the drift problem to be not well solved in realistic scenarios. In this study, we propose a new urban time series prediction model for the concept drift problem, which encodes the drift by considering the periodicity in the data and makes on-the-fly adjustments to the model based on the drift using a meta-dynamic network. Experiments on real-world datasets show that our design significantly outperforms state-of-the-art methods and can be well generalized to existing prediction backbones by reducing their sensitivity to distribution changes. 
\end{abstract}



\maketitle

\begin{figure}[h]
	\centering
	\includegraphics[width=0.45\textwidth]{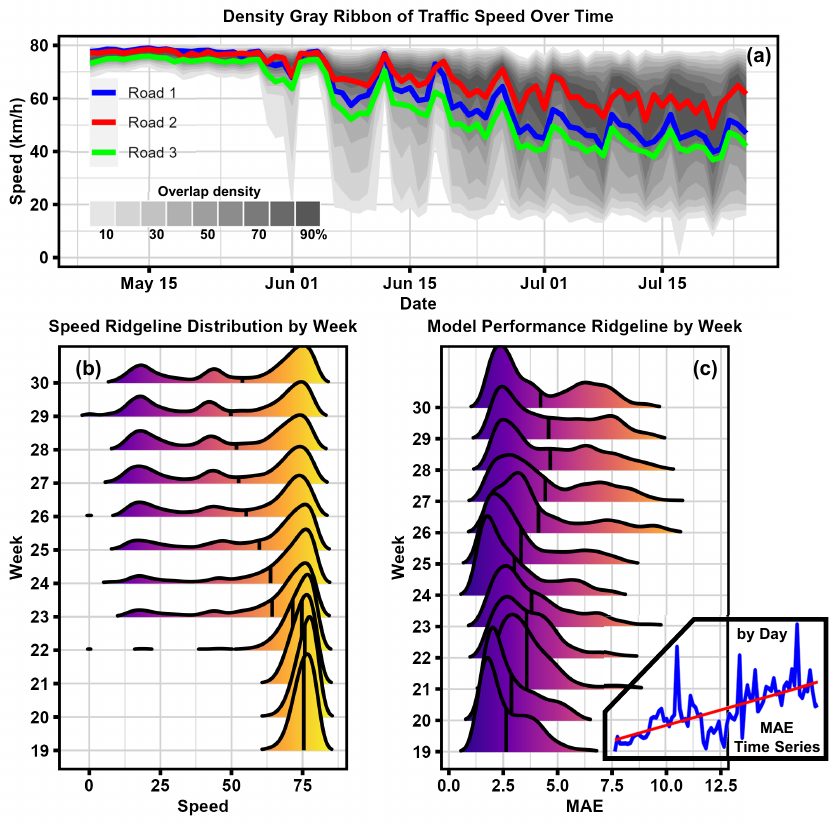}
	\caption{Illustration of traffic speed concept drift in Beijing from May 8 to July 25, 2022. (a) Grayscale ribbon of road speeds, with highlighted data for three sampled roads. (b) The ridgeline plot exhibits weekly changes in the road speed distribution. (c) The ridgeline plot exhibits weekly changes in the error distribution of the predictive model, and the subplot shows the trend of the daily error.}
	\label{fig:concept_drift}
\end{figure}

\section{Introduction}
Accurate and reliable forecasting of urban multivariate time series data, such as traffic speed, travel demand, and electricity consumption, has been intensively studied as an essential task since it contributes to reducing traffic congestion, energy waste, and greenhouse gas emissions \cite{ruan2020dynamic,jia2017spatial,yao2018deep,jiang2019deepurbanevent,jiang2023spatio}. The advancement of machine learning, particularly deep learning \cite{vaswani2017attention,li2018diffusion}, has facilitated both research and practical applications in this field. However, the underlying assumption that these algorithms rest on - independent and identical distributions (i.i.d.) - may not always hold true in the fast-paced, dynamic urban environment teeming with implicit event-induced uncertainty. As an illustrative example, consider the changes in daily traffic speeds in Beijing as shown in Fig. \ref{fig:concept_drift}. The grayscale ribbon plot in Fig. \ref{fig:concept_drift}(a) demonstrates that the road speed decreases over Mid-2022, indicating changes in the road speed distribution. The evolution of this distribution, as captured week-by-week in the ridgeline plots of Fig. \ref{fig:concept_drift}(b), shows the shift from a concentration around 75 km/h to a broader distribution around 10-75 km/h. The drift distribution imposes vulnerability on models trained from earlier data as they are not expected to generalize well to the new coming data, known as the Concept Drift problem \cite{widmer1996learning}. As a result, Fig. \ref{fig:concept_drift}(c) highlights an increase in the error of a pre-trained model over time, with error distribution spreading from low to medium and high. These changes reflect the realities of a city in flux, where social, economic, and environmental factors can drastically affect urban patterns, leading to concept drift. It is considered a primary reason for the declining effectiveness of intelligent information systems, as the change in data distribution weakens the induced correlation patterns between new and past data \cite{lu2018learning}. Addressing this challenge can enhance the equity, safety, and sustainability of cities and communities, assisting urban decision-makers and stakeholders in using data-driven approaches to tackle urban problems amidst continuous social changes.

Learning from changing data has attracted considerable research interest. For non-stationary time series with concept drift, studies are conducted to: (1) detect the occurrence of concept drift by error rate-based \cite{gama2004learning,bifet2007learning} or data distribution-based \cite{kifer2004detecting,bhatia2022memstream} detectors; (2) train a new model, fine-tune the current model, or ensemble the new and existing model \cite{li2022ddg,du2021adarnn,you2021learning} to fit the latest data. However, such incremental learning approaches can only adjust the model to narrow the gap among the drifting distributions. Awaiting the accumulation of new data inevitably leads to lagging. The latest adapted model could fail again if the drift continues. And even many scenarios do not qualify for frequent retraining as periodic updates are costly. Considering that urban models are often associated with substantial economic benefits and sustainable social development, this prompts us to explore an evolving approach beyond retraining that can be proactively adjusted on-the-fly based on inputs.

Nevertheless, developing a self-configuring learner is non-trivial. A crucial consideration is the computational efficiency. Addressing concept drift implies the necessity for the model to process extensive historical data, providing a comprehensive understanding of past data distribution to accurately perceive ongoing drifts. However, the assimilation and retention of large volumes of historical data impose a significant computational burden. The model must therefore exhibit the capacity to manage historical data in an intelligent manner, maintaining long-term historical data whilst picking the most valuable information. Second, the navigation of the stability-plasticity dilemma cannot be overlooked. An adaptive model with essential levels of responsiveness to new trends must achieve an equilibrium between maintaining prior learned knowledge (stability) and accommodating new data patterns (plasticity). This poses high requirements in deep learning models as the parameter space of the model can be extremely large. Searching this space for the optimal configuration involves considerable degrees of freedom and can be practically infeasible for large models.

To confront the identified challenges, we propose a novel drift adaptation network designed around two key components: a dual memory module and a strategically adjustable meta-dynamic network. Firstly, our model incorporates a dual memory module to enhance computational efficiency while handling extensive historical information. This module consists of a Replay Memory for time-efficient drift embedding and a Pattern Memory for preserving significant patterns over extended periods. Then a novel drift-adaptive meta-dynamic network is utilized to generate adjustable parameters for bridging the encoder and decoder, allowing the model to respond to changing data patterns with minimal degrees of freedom. Our approach is both simple and general, as it only requires tracking a few parameters to model the drift and can be highly compatible with any modern urban deep learning models as a plug-and-play module for handling concept drift. Compared to existing models, the new network improves prediction under non-stationary time series in a robust and self-generating manner. Our contribution is summarized in the following sections.

\begin{itemize}
    \item We first propose to model the urban concept drift by finding learnable components in non-stationarity, which facilitates a more nuanced understanding of urban dynamics and introduces new perspectives for prediction in such environments.
    \item We propose MemDA, an on-the-fly adaptation architecture that implicitly perceives drift without retraining and dynamically adjusts the corresponding model parameters. 
    \item The proposed model is pluggable and backbone-agnostic. Extensive experiments on four urban datasets demonstrate the superior performance of our design\footnote{\url{https://github.com/deepkashiwa20/Urban_Concept_Drift}}.
\end{itemize}

\section{Related Work}
\subsection{Concept Drift and Model Adaptation}
Time-varying patterns in the data render obsolete the knowledge previously gained from old and static data, thereby reducing the predictive capability of the model for new data \cite{lu2018learning,khamassi2018discussion}. As the gap between the latest and upcoming data distribution is more likely narrow, the fresh training model can be utilized as a tool to address the issue. There are three types of approaches based on trade-offs for preserving historical information: the straightforward way is to retrain a new model with the most recent data. An explicit sliding window drift detector is commonly employed to determine when to retrain \cite{bifet2007learning,li2022ddg}. Alternatively, in some cases, retaining and fine-tuning the old model can save efforts in training the new model. New base predictors can be added to existing collections of models, or data-adaptive learning models be created. \cite{you2021learning} designs multiple predictors using a deep ensemble approach and trains a classifier to predict which predictor to use. \cite{zhao2020handling} proposes a weighted method for reusing previous models, where each model is associated with a metric that represents its reusability for the current observation. \cite{du2021adarnn} uses a segmentation algorithm to divide the time series and uses a transfer learning approach to integrate knowledge under various concepts. However, training-based approaches are just attempting to catch up with the evolution of the concept. They can only build models after drift and inevitably suffer from model aging.

Normalizing the input into a normalized distribution is also used to deal with non-stationarity. RevIN\cite{kim2021reversible} to remove and rebuild the discrepancy in the mean and standard deviation of the input sequences. The Non-stationary Transformer\cite{liu2022non} mitigates the over-stationarization problem by calculating the normalized attention. Dish-TS\cite{fan2023dish} learns sample-level normalization parameters. However, The normalization-based approaches simply attempt to align the magnitudes of the samples, with a weak capacity for modeling changes in data patterns and conditional probability distributions.

\subsection{Urban Time Series Forecasting}
Building time series models to forecast future traffic speed, taxi demand, electricity consumption, crowd in/outflow, etc., are widely studied as key technologies for smart city applications. \cite{yao2018deep} and \cite{cui2019traffic} employ an LSTM structure to connect hidden embeddings for urban data forecasting. \cite{lai2018modeling}, \cite{li2018diffusion}, \cite{bai2020adaptive}, and \cite{zhao2019t} use GRU for more efficient modeling of traffic volume and speed.  Meanwhile, convolution-based modules, such as 1D CNN and TCN, are shown to be competent with temporal data as well. \cite{yu2018spatio}, \cite{shih2019temporal}, \cite{wu2019graph}, and \cite{wu2020connecting} for predicting urban data demonstrate the efficiency and effectiveness of such designs. Attention mechanisms in natural language processing have also been adapted. \cite{yao2019revisiting,guo2019attention} design models with multiple inputs based on the periodicity of the data (hours, days, or weeks) and subsequently use attention to capture the dynamic temporal correlation and assign greater weight to important moments. \cite{zheng2020gman} and \cite{xu2020spatial} also propose similar attention modules for better global/long-term temporal modeling. However, mainstream efforts in urban data forecasting assume constant data patterns. The pre-trained models are highly vulnerable to the drift in data.

Another line focuses on the prediction of urban data under events. \cite{song2014prediction} predicts human movements during earthquakes using a Hidden Markov Model. \cite{jiang2019deepurbanevent,jiang2023learning} models the crowd dynamics during four big events by mobility momentum; EAST-Net \cite{wang2022event} enhances the robustness of the deep model using heterogeneous information networks and memory bank. However, prediction in the event case is closer to emergency response, which is concerned with the effect of the model on localized outliers, whereas the concept drift is focused on persistent changes in the patterns underlying the data.

\section{Preliminaries}
\textbf{Concept Drift.} Concept drift refers to the underlying distribution of streaming data varies over time caused by changes in unobserved hidden variables \cite{lu2018learning}. It will lead to model performance degradation as the statistical properties of the target domain do not match the history. Formally, given a set of instances obtained at different timesteps $\{(x_0, y_0),..., (x_T, y_T)\}$, where $x_{t}$ is the feature and $y_{t}$ is the target, concept drift at timestamp $t$ can be described as $p_{0:t-1}(x,y)\ne p_{t:T}(x,y)$, where $p(\cdot)$ denotes the joint probability distribution. The source of concept drift can be decomposed into two parts: changes in the marginal probability distribution $p(x)$ and changes in the conditional probability distribution $p(y|x)$.

\noindent\textbf{Adaptive Forecasting.} Given a specified granularity of urban time series data, the tensor $x_{t}\in \mathbb{R}^{N\times C}$ can be represented as the observation at timestamp $t$, where $N$ denotes the number of urban sensors (or regions) and C denotes the feature channel (e.g., traffic speed, crowd). Then given the historical observations $\{x_{i}|i\in[0,t]\}$, building adaptive model for the future $\alpha$ steps of urban data $\hat{Y}_{t} = \{\hat{x_{i}}|i\in[t+1,t+\alpha]\}$ is to obtain such parameters $\theta_{t}$ that can minimize the objective function $\mathcal{L(\cdot)}$ as follows:
\begin{equation}
\theta_{t} = \mathop{\arg\!min}_{\theta_{t}}{\mathcal{L}(\hat{Y}_{t},\{x_{i}|i\in[t+1,t+\alpha]\})}.
\end{equation}
where $x_{i}$ are drawn from a time-varying distribution. $\theta_{t}$ is learned from the training data and will be continuously adapted to test data.

\begin{table}[h]
	\centering
	\caption{Notations}\label{tab:notation}		
    \resizebox{.45\textwidth}{!}{
	\begin{tabular*}{9.8cm}{@{\extracolsep{\fill}}ll}
		\hline
		\multirow{1}{0.6cm}{Symbol} &
		\multicolumn{1}{c}{Description}
		\\
		\hline
        \hline
        $x_{t}$             & Urban data observation tensor at timestamp $t$ \\
        $N, C$              & Dimension of $x_{t}$ \\
        $T$                 & Total number of training samples\\
        $p(x,y)$            & Joint probability distribution of variables $x$ and $y$ \\
        $\alpha$            & Prediction leading time \\
        $\mathit{X}_t$      & Observations of urban data over time $[t-\alpha,t]$ \\
        $\mathcal{X}_t$     & A collection of observations constructed by periodicity\\
        $p$                 & Data sampling volume in one day \\
        $k$                 & the k-th look-back day for drift embedding \\
        $P_k$               & Look-back timestamp for drift embedding, $P_k = p\times k$ \\
        $K$                 & Total look-back days \\
        $\mathcal{F}$       & Neural network functions \\
        $\mathit{Z}_t$      & Embedding result of $\mathit{X}_t$  by Encoder\\
        $C_e$               & Dimension of embedding generated by encoder \\
        $\mathit{E}_t$      & Embedding result of $\mathcal{X}_t$ by Encoder and Replay Memory\\
        $\mathit{M}$        & Pattern Memory \\
        $L, D$              & Dimension of Pattern Memory \\
        $\mathit{V}_t$      & The query result of $\mathit{E}_t$ returned by Pattern Memory \\
        $H_t$               & Drift Embedding result of $\mathcal{X}_t$ by Dual-Memory\\
        $W_t$               & Drift adaptation parameters\\
        $N_s$               & Dimension of similarity generated by NTN\\
        $I(X;Y)$            & Mutual information between variables $X$ and $Y$ \\
		\hline
	\end{tabular*}}
\end{table} 

\begin{figure*}[h]
	\centering
	\includegraphics[width=0.95\textwidth]{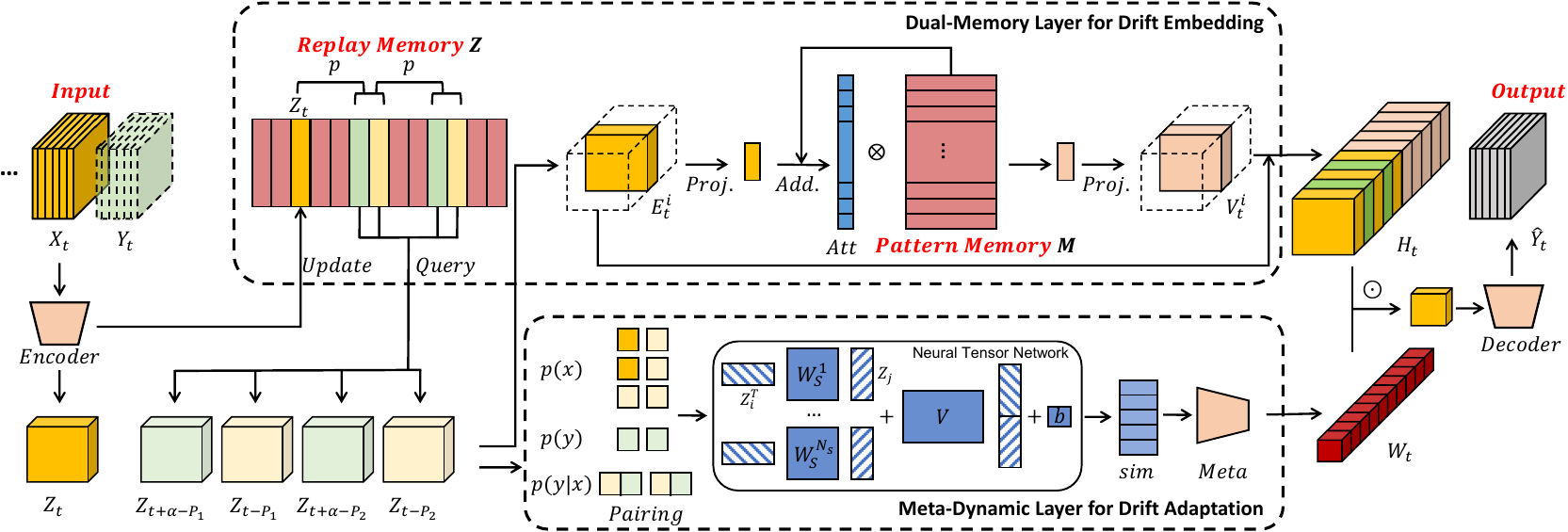}
	\caption{The framework of the memory-based drift adaptation model.}
	\label{fig:framework}
\end{figure*}

\section{Methodology}
\subsection{Rationale}
Although the distribution of urban data may change due to a variety of factors, the inherent periodicity of urban dynamics follows a consistent temporal regularity\cite{gonzalez2008understanding}. This persistent feature enables the model to maintain its predictive capabilities, as the accumulation and integration of fresh patterns facilitate adaptive learning even when distribution undergoes significant transformations. It serves as an anchor for our model, guiding the model toward the interpretation and management of concept drift. Moreover, we acknowledge the sustained features to be encoded in the pre- and post-drift data retain significant commonalities. These sustained features can be efficiently captured by the stable component of the model, which means modifications to some key model parameters should be prioritized instead of updating all parameters. This strategy strikes a balance between stability and plasticity, ensuring the resilience of the model to varying data patterns and distributions.

Motivated by these, we design a \textbf{Mem}ory-based \textbf{D}rift \textbf{A}daptation (\textbf{MemDA}) network illustrated in Fig.~\ref{fig:framework}, where we first propose to use a Dual-Memory to encode periodicity implied in the long sequential inputs to counteract the drift, and later use meta dynamic networks to generate adaptation parameters governing the integration of input information. The notations are listed in Table \ref{tab:notation}.

\subsection{Dual-Memory for Drift Embedding}
\begin{figure}[h]
	\centering
	\includegraphics[width=0.46\textwidth]{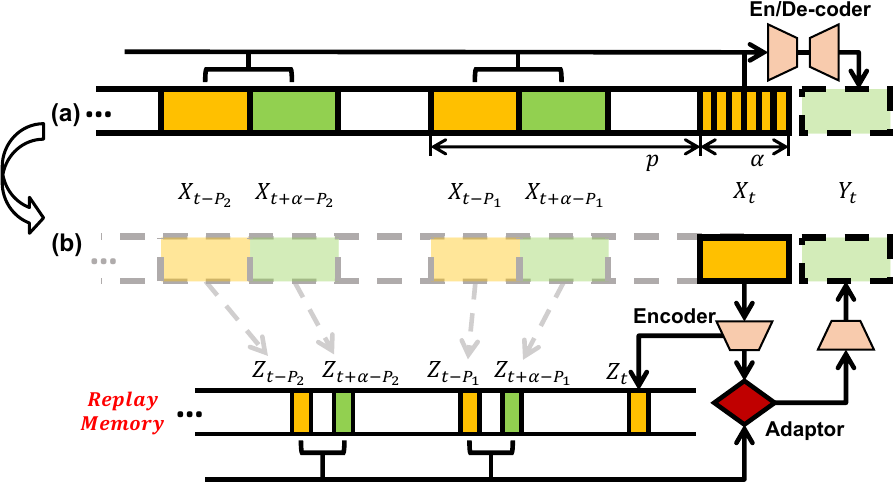}
	\caption{Design of model inputs switches from Plain Blocks (a) to Replay Memory enhanced (b) inputs.}
	\label{fig:input_structure}
\end{figure}

Periodic inputs are essential for coping with urban data dynamics, while long sequential inputs are the informational foundation for perception to drift. Therefore, as shown in Fig. \ref{fig:input_structure}(a), we accordingly construct \textbf{Drift-Aware Inputs} $\mathcal{X}_{t}$ along the time axis. Assuming that the data are sampled at a frequency of $p$ per day, the inputs are constructed by fetching up to $K$ days of data as follows:
\begin{align}
    \mathcal{X}_{t} &= \{\mathit{X}_t\} \cup \{\mathit{X}_{t-P_k}, \mathit{X}_{t+\alpha-P_k} | k \in [1, K]\}, \nonumber \\
    \intertext{where $P_k = p\times k$ and $\mathit{X}_t$, $\mathit{X}_{t-P_k}$, $\mathit{X}_{t+\alpha-P_k}$ are:}
    \mathit{X}_t &= \{x_{t-\alpha}, x_{t-\alpha+1}, ..., x_{t}\}, \nonumber \\
    \mathit{X}_{t-P_k} &= \{x_{t-\alpha-P_k}, x_{t-\alpha+1-P_k}, ..., x_{t-P_k}\}, \nonumber \\
    \mathit{X}_{t+\alpha-P_k} &= \{x_{t+1-P_k}, x_{t+2-P_k}, ..., x_{t+\alpha-P_k}\}.
\end{align}
The segment $\mathit{X}_t$ consists of the closest $\alpha$ observations up to time $t$. The segments $\mathit{X}_{t-P_k}$ and $\mathit{X}_{t+\alpha-P_k}$ consist of the same number of observations as $\mathit{X}_t$ but from before and after the same clock $t$ prior to $k$ days. Here, $P_k$ represents the time gap between the backtrack and the present. Thus, $\mathcal{X}_{t}$ contains $2K+1$ segments and  each segment has $\alpha$ observations.

However, for such plain long time series (i.e., $(2K+1)\times\alpha$ steps), traditional approaches~\cite{zhang2017deep,guo2019attention,jiang2021deepcrowd} that utilize multi-branch to encode the sequence will be no longer applicable as the computational complexity exponentially increases as the input grows. Therefore, we propose to use a \textbf{Replay Memory (RM)} for efficient drift embedding. RM is essentially a queue that stores all the hidden embedding of $\mathit{X}_t$ computed by the encoder, forming $\mathit{Z}\in\mathbb{R}^{T\times N\times C_e}$, where $T$ is the number of training samples and $C_e$ is the output dimension of the encoder. During training, the RM fetches the embeddings $\mathit{Z}_{t-P_k}$ and $\mathit{Z}_{t+\alpha-P_k}$ of the past segments $\mathit{X}_{t-P_k}$ and $\mathit{X}_{t+\alpha-P_k}$ to avoid duplicate computing. Without loss of generality, we use $\mathcal{F}_e$ to denote the encoder network with parameter $\theta_e$. As shown in Fig. \ref{fig:input_structure}(b), the embedding result $\mathit{E}_t$ for Drift-Aware Inputs $\mathcal{X}_t$ is converted to the $\mathit{X}_t$ encoding and RM replaying as:

\begin{equation}
    \mathit{E}_t = \underbrace{\{\mathcal{F}_e(\mathit{X}_t, \theta_e)\}}_{\text{Encoding}} \cup \underbrace{\{\mathit{Z}_{t-P_k}, \mathit{Z}_{t+\alpha-P_k}|k\in[1,K]\}}_{\text{Replaying}}.
\label{eq:et}
\end{equation}
The embedding of $\mathit{X}_t$ is subsequently updated to RM by $\mathit{Z}_t = \mathcal{F}_e(\mathit{X}_t, \theta_e)$. Note that replayed memories are collected from the previous epoch while $\mathit{Z}_t$ is obtained at the current epoch (i.e., one epoch lag) when training. However, as the epoch proceeds, they will gradually converge to the same embedding space.

While periodic long inputs ensure the possibility of tracking data drift, the model cannot extend segments indefinitely, and distribution information may not be limited to periodicity. Therefore, we further augment the drift embedding $E_t$ with a \textbf{Pattern Memory (PM)}. The PM explicitly records the prototype of normal patterns during training and serves to perceive out-of-distribution during inference. Specifically, the Pattern Memory $M\in \mathbb{R}^{L\times D}$ is a parameter matrix with $L$ real-valued $D$ dimensional prototypes. For the $i$-th embedding in $\mathit{E}_t$, $M$ is queried to find similar prototypes by attention. Formally, we have:
\begin{equation}
\left\{
\begin{aligned}
    Q_i &= E_t^i\mathit{W}_Q+b_Q \\
    \varphi_j &= \frac{e^{Q_i\ast M_j}}{\sum_{j=1}^{L}e^{Q_i\ast M_j}} \\
    \mathit{V}_t^i &= (\sum_{j=1}^{L}\varphi_j\cdot M_j)\mathit{W}_V+b_V,
\label{eq:att}
\end{aligned}
\right.
\end{equation}
where $\varphi_j$ is the attention score corresponding to $j$-th prototype; $\mathit{W_Q}$, $b_Q$, $\mathit{W_V}$, $b_V$ are linear transformation parameters; $\mathit{V}_t^i$ denotes the query result of $\mathit{E}_t^i$.

Then, $\mathit{E}_t$ and $\mathit{V}_t$ are concatenated to form the final drift embedding $\mathit{H}_t=[\mathit{E}_t,\mathit{V}_t]$, which encodes the changes in data distribution and implies the essential information for prediction. In general, the weighted fusion of embedding can integrate all the information and can be fed to the decoder for future predictions. One mainstream approach \cite{zhang2017deep,yao2019revisiting,bai2020adaptive} is to use the trainable parameter to estimate the influence weights of each segment as: 
\begin{equation}
    \hat{Y}_t = \mathcal{F}_{proj.}(\sum_{i=1}^{4K+2}W^i\odot \mathit{H}_t^i)
\label{eq:fusion}
\end{equation}
where $\odot$ is Hadamard product; $W\in\mathbb{R}^{4K+2}$ are learnable parameters, and $\mathcal{F}_{proj.}$ projects the result to the prediction. Next, we introduce a dynamic network to make the key parameters $W$ self-configurable.

\subsection{Meta-Dynamic for Drift Adaptation}
We characterize model adaptation in terms of Information Bottlenecks (IB). The IB problem was introduced in \cite{tishby2000information} as an information-theoretic framework for learning. It considers quantifying the amount of information of hidden representation $H$ extracted from the observation $X$ contains about another relevant target signal $Y$ by using Mutual Information (MI). The goal is to find a $H$ that minimizes the mutual information $I(X;H)$ while keeping $I(H;Y)$ above the threshold. IB theory suggests that deep neural networks construct Markov chains of the hidden layer representations about the input $X$, and DNNs trained using SGD are essentially solving the compression-prediction trade-off problem \cite{goldfeld2020information}, which means deeper layers correspond to smaller $I(X;H)$ but larger $I(H;Y)$.

When it comes to the concept drift problem, in the encoder-decoder network, the encoder essentially acts as an information bottleneck that filters the noise from the data and forces the network to extract the typical patterns of high-dimensional data, of which outputs are then transmitted to the decoder. This makes $I(X;H)$ and $I(H;Y)$ a function of the en/decoder parameters. When drift occurs, the encoder is more resistant to drift as it is more focused on information compression. This leads us to fix the encoder parameters $\theta_e$ to reduce the degrees of freedom while preserving the ability to extract features over different target domains. For the adaptor, since we want to boost the MI between the hidden embedding $\mathit{H}_t$ and $Y_t$, the fusion parameter $W$ should be actively and dynamically adjusted to different magnitudes under different concepts, i.e., varying with time $t$. For example, the weight of historical information is evenly distributed when the data is stable, while the influence of history is reduced when the data is out of distribution. To generate the appropriate $W_t$, it is necessary to decide how much each segment embedding contributes to the future, which means it should be proportional to the MI of $\mathit{H}_t$ and $Y_t$, i.e.,
\begin{equation}
\begin{aligned}
    W_t &= softmax(I(\mathit{H}_t^1;Y_t), ..., I(\mathit{H}_t^{4K+2};Y_t))\\
          &= softmax(\mathbb{E}_{Y_t}\{D_{KL}(p(\mathit{H}_t^1)\|p(\mathit{H}_t^1|Y_t)), ...,\\ &D_{KL}(p(\mathit{H}_t^{4K+2})\|p(\mathit{H}_t^{4K+2}|Y_t))\}).
\end{aligned}
\label{eq:MI}
\end{equation}

However, since the computation of MI requires knowledge of the joint and marginal distributions of two random variables, the distributions that can be depicted for high-dimensional urban datasets are stretched compared to the true continuous distribution, especially when the joint distribution changes in drift scenarios. This makes it tough to compute MI directly. Nevertheless, it is possible to approximate the MI between high-dimensional variables using a neural network with gradient descent optimization\cite{belghazi2018mine}. Therefore, we propose to use a meta-dynamic network to approximate the contribution of each segment by dynamically generating the corresponding $W_t^i$ conditional on the input. Specifically, to perceive the drift, we construct three lists of temporal alignment tuples,
\begin{equation}
\begin{aligned}
    Pair_x &= \{(\mathit{Z}_t,\mathit{Z}_{t-P_1})\}\cup\{(\mathit{Z}_{t-P_k},\mathit{Z}_{t-P_{k+1}})\}, \\
    Pair_y &= \{(\mathit{Z}_{t+\alpha-P_k},\mathit{Z}_{t+\alpha-P_{k+1}})\}, \\
    Pair_{xy} &= \{([\mathit{Z}_{t-P_k}, \mathit{Z}_{t+\alpha-P_{k}}], [\mathit{Z}_{t-P_{k+1}}, \mathit{Z}_{t+\alpha-P_{k+1}}])\}. \\
\end{aligned}
\end{equation}
The constructed embedding pairs are used to measure the evolution of the distribution over the marginal probability $p(x)$ and the conditional probability $p(y|x)$. Each pair serves as an anchor to facilitate determining the contribution of each segment embedding to the prediction. Each $(\mathit{Z}_i,\mathit{Z}_j)\in Pair$ is fed into a Neural Tensor Network (NTN) \cite{socher2013reasoning} to model the relation:
\begin{equation}
\begin{aligned}
    sim_k & = \sigma(\mathit{Z}_{i}^{T}W_S\mathit{Z}_{j} + V[\mathit{Z}_{i}, \mathit{Z}_{j}]^{T} + b_S),\\
    W_t &= \mathcal{F}_{meta}([sim_1, sim_2, ...]),
\end{aligned}
\label{eq:adapW}
\end{equation}
where $W_S\in\mathbb{R}^{C_e\times C_e\times N_s}$, $V\in\mathbb{R}^{N_s\times 2C_e}$, and $b_S\in \mathbb{R}^{N_s}$ are weight tensor. $N_s$ is a hyperparameter that controls the drift scores generated by the model for each pair from different perspectives. $\sigma$ is an activation function. $\mathcal{F}_{meta}$ denotes a meta layer that can be implemented in various ways, which generates fusion parameters based on the similarity matrix to adjust to the drift on-the-fly. The generated $W_t$ are loaded into the fusion layer replacing the static $W$ in Eq.~\ref{eq:fusion}, and a multilayer CNN is employed as a decoder to project the result to the prediction. The groundtruth values are taken to calculate the loss with prediction to optimize the model. 

\subsection{Training Details}
During the training phase, we train the model by random sampling batch, and the RM is updated between epochs. During the testing phase, we feed the samples into the model in chronological order, and the RM is updated sequentially. Note that in the training phase, we maintain a full number of embeddings in RM as many as $T$, while in the testing phase and online environment, we only retain the most recent $p\times K$ embeddings to prevent unlimited growth. The PM is updated as the gradient descends and is fixed in the test.

\section{Experiments}

\begin{table}[h]
    \centering
	\caption{Summary of Experimental Datasets}
	\label{tab:data}
    \resizebox{.48\textwidth}{!}{
	\begin{tabular*}{10.8cm}{@{\extracolsep{\fill}}lcccc}
        \hline
        \hline
		\textbf{Dataset} & Period & \#$N$ & Temporal & Drift\\
		\hline
        \textbf{PeMS} & 2020/01/01$\sim$2020/07/31  & 325 & 5 minutes & Sudden \\
        \hline
        \textbf{Beijing} & 2022/05/12$\sim$2022/07/25 & 3126 & 5 minutes & Sudden \\
        \hline
        \textbf{Electricity} & 2012/01/01$\sim$2012/06/30 & 370 & 1 hour & Incremental \\
        \hline
        \textbf{COVID-CHI} & 2019/07/01$\sim$2020/12/31 & 112 & 2 hour & Incremental \\
        \hline
        \hline
	\end{tabular*}}
\end{table}

\subsection{Datasets}
\begin{figure}[h]
	\centering
	\includegraphics[width=0.48\textwidth]{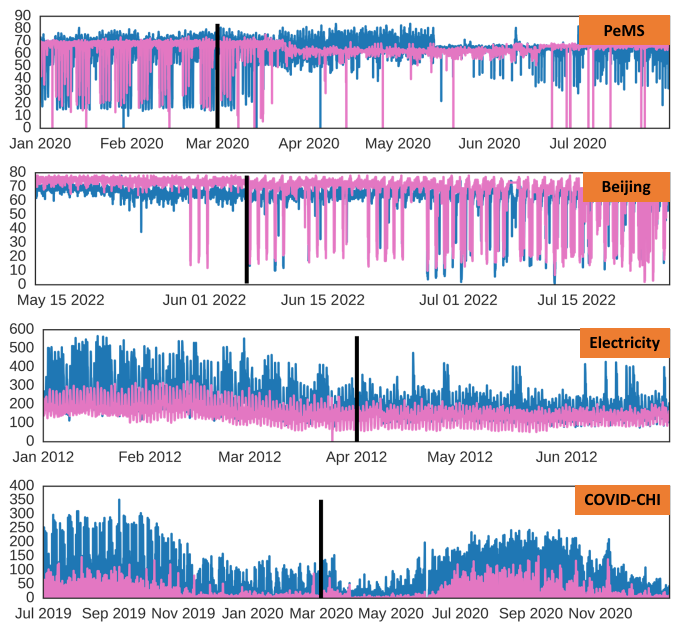}
	\caption{Two nodes from each of the four datasets are drawn for time series plots. The black vertical lines indicate the cut-off positions of the training and testing datasets.}
	\label{fig:data}
\end{figure}

We collected data from four cities covering different data sources and concept drift types, which are presented in Table \ref{tab:data}. The PeMS and Beijing datasets are collected from the traffic speeds of major roads in California and Beijing. The Electricity dataset contains the electricity consumption. And the COVID-CHI \cite{wang2022event} dataset is the demand for shared bicycles collected from Chicago. To better understand how the data distribution has changed, we show sampled data in Fig. \ref{fig:data}, where each dataset exhibits different types of concept drift. These drifts are caused by known events (i.e., COVID, Seasonality) and result in significant variations in the data distribution.

\begin{table*}[h]
        \footnotesize
	\centering
	\caption{Performance Comparisons with state-of-the-art Urban Time Series Prediction Models}
	\label{tab:mse}
    \resizebox{.98\textwidth}{!}{
	\begin{tabular*}{15.2cm}{@{\extracolsep{\fill}}c||ccc|ccc|ccc|ccc}
		\hline
        \multirow{2}{1.0cm}{\textbf{Model}}&
		\multicolumn{3}{c|}{\textbf{PeMS}} &
        \multicolumn{3}{c|}{\textbf{Beijing}} &
        \multicolumn{3}{c|}{\textbf{Electricity}} &
		\multicolumn{3}{c}{\textbf{COVID-CHI}}
		\\
		\cline{2-13}
        \multicolumn{1}{c||}{} & 
		\multicolumn{1}{c}{\textit{RMSE}} & 
		\multicolumn{1}{c}{\textit{MAE}} &
		\multicolumn{1}{c|}{\textit{MAPE}} & 
        \multicolumn{1}{c}{\textit{RMSE}} & 
		\multicolumn{1}{c}{\textit{MAE}} &
		\multicolumn{1}{c|}{\textit{MAPE}} & 
        \multicolumn{1}{c}{\textit{RMSE}} & 
		\multicolumn{1}{c}{\textit{MAE}} &
		\multicolumn{1}{c|}{\textit{MAPE}} & 
		\multicolumn{1}{c}{\textit{RMSE}} & 
		\multicolumn{1}{c}{\textit{MAE}} & 
		\multicolumn{1}{c}{\textit{MAPE}}
		\\
		\hline
        CLD & 4.022 & 1.784 & 3.140\% & 11.028 & 5.225 & 15.029\% & 81.325 & 41.088 & 15.422\% & 15.879 & \underline{6.789} & 86.680\% \\
        ARIMA & 3.135 & 1.215 & 2.301\% & 9.184 & 4.640 & 16.400\% & 94.334 & 64.808 & 18.222\% & 16.721 & 6.903 & 80.334\% \\
        AGCRN & 2.572 & 1.191 & 2.094\% & 8.715 & 4.238 & 14.376\% & 95.242 & 50.422 & 15.451\% & 16.279 & 7.087 & 85.224\% \\
        DCRNN & 2.500 & 1.158 & 2.030\% & 8.280 & 3.811 & 13.838\% & 125.335 & 125.335 & 25.754\% & 17.453 & 7.175 & 87.617\% \\
        MTGNN & 2.426 & 1.148 & 2.007\% & 10.976 & 5.120 & 19.928\% & 96.475 & 51.562 & 15.410\% & 18.721 & 7.434 & 89.097\% \\
        STGCN & 2.556 & 1.212 & 2.114\% & 10.284 & 5.000 & 15.284\% & 132.209 & 72.802 & 22.580\% & 22.462 & 8.687 & 85.064\% \\
        ASTGCN & 2.402 & 1.157 & 2.078\% & 9.156 & 4.250 & 17.130\% & \underline{74.074} & \underline{39.388} & \underline{13.584\%} & 15.242 & 6.830 & \underline{77.223}\% \\
        GMAN & 2.849 & 1.298 & 2.308\% & 11.555 & 6.448 & 21.941\% & 138.436 & 81.285 & 26.263\% & 22.462 & 8.687 & 85.064\% \\
        StemGNN & 2.414 & 1.144 & 2.030\% & 8.575 & 4.294 & 15.605\% & 103.375 & 56.803 & 19.366\% & 17.231 & 7.132 & 85.32\% \\
        GW-Net & \underline{2.385} & 1.125 & \underline{1.976\%} & \underline{7.754} & \underline{3.564} & \underline{11.916\%} & 97.087 & 51.684 & 16.120\% & 16.210 & 6.857 & 80.038\% \\
        STTN & 2.441 & 1.168 & 2.040\% & 9.886 & 4.748 & 17.150\% & 93.125 & 50.671 & 15.776\% & \underline{14.853} & 6.934 & 79.155\% \\
        EAST-Net & 2.399 & \underline{1.118} & 1.988\% & 11.671 & 5.970 & 21.119\% & 100.789 & 55.158 & 18.714\% & 17.543 & 7.832 & 82.321\% \\
        \hline
        \textbf{MemDA} & \textbf{2.297} & \textbf{1.053} & \textbf{1.861\%} & \textbf{6.720} & \textbf{3.192} & \textbf{9.913\%} & \textbf{67.413} & \textbf{34.814} & \textbf{12.186\%} & \textbf{14.003} & \textbf{6.115} & \textbf{72.088\%} \\
        $\Delta$\% & 3.70\% & 5.82\% & 5.78\% & 13.34\% & 10.43\% & 16.81\% & 8.99\% & 11.61\% & 10.29\% & 5.73\% & 9.93\% & 9.93\% \\
		\hline
	\end{tabular*}}
\end{table*}

\begin{table*}[h]
	\centering
	\caption{Performance Comparisons with state-of-the-art Adaptive Models}
	\label{tab:adap_mse}
    \resizebox{.98\textwidth}{!}{
	\begin{tabular*}{19.2cm}{@{\extracolsep{\fill}}c||ccc|ccc|ccc|ccc}
		\hline
        \multirow{2}{1.0cm}{\textbf{Model}}&
		\multicolumn{3}{c|}{\textbf{PeMS}} &
        \multicolumn{3}{c|}{\textbf{Beijing}} &
        \multicolumn{3}{c|}{\textbf{Electricity}} &
		\multicolumn{3}{c}{\textbf{COVID-CHI}}
		\\
		\cline{2-13}
        \multicolumn{1}{c||}{} & 
		\multicolumn{1}{c}{\textit{RMSE}} & 
		\multicolumn{1}{c}{\textit{MAE}} &
		\multicolumn{1}{c|}{\textit{MAPE}} & 
        \multicolumn{1}{c}{\textit{RMSE}} & 
		\multicolumn{1}{c}{\textit{MAE}} &
		\multicolumn{1}{c|}{\textit{MAPE}} & 
        \multicolumn{1}{c}{\textit{RMSE}} & 
		\multicolumn{1}{c}{\textit{MAE}} &
		\multicolumn{1}{c|}{\textit{MAPE}} & 
		\multicolumn{1}{c}{\textit{RMSE}} & 
		\multicolumn{1}{c}{\textit{MAE}} & 
		\multicolumn{1}{c}{\textit{MAPE}}
		\\
		\hline
        Backbone & 2.385 & 1.125 & 1.976\% & 7.754 & 3.564 & 11.916\% & 97.087 & 51.684 & 16.120\% & 16.210 & 6.857 & 80.038\% \\
        \hline
        RevIN & 2.541 & 1.129 & 1.956\% & 8.116 & 3.707 & 11.200\% & 85.352 & 45.434 & 15.847\% & 15.472 & 6.789 & 80.049\% \\
        $\Delta$\% & \multicolumn{3}{c|}{No Improvement} & \multicolumn{3}{c|}{No Improvement} & 12.1\% & 12.1\% & 1.7\% & 4.6\% & 1.0\% & - \\
        \hline
        Dish-TS & 2.519 & 1.144 & 1.973\% & 7.738 & 3.586 & 10.499\% & 73.509 & 41.465 & 13.726\% & 14.807 & 6.435 & 74.182\% \\
        $\Delta$\% & \multicolumn{3}{c|}{No Improvement} & \multicolumn{3}{c|}{No Improvement} & 24.3\% & 19.8\% & 14.8\% & 14.8\% & 12.7\% & 14.0\% \\
        \hline
        MemDA & \textbf{2.297} & \textbf{1.053} & \textbf{1.861}\% & \textbf{6.720} & \textbf{3.192} & \textbf{9.913\%} & \textbf{67.413} & \textbf{34.814} & \textbf{12.186\%} & \textbf{14.003} & \textbf{6.115} & \textbf{72.088}\% \\
        $\Delta$\% & 3.7\% & 6.4\% & 5.8\% & 13.3\% & 10.4\% & 16.8\% & 30.6\% & 32.6\% & 24.4\% & 19.4\% & 17.1\% & 16.4\% \\
		\hline
	\end{tabular*}}
\end{table*}

\begin{table*}[h]
	\centering
	\caption{Variants Performance Evaluation}
	\label{tab:ablation}
    \resizebox{.98\textwidth}{!}{
	\begin{tabular*}{18.2cm}{@{\extracolsep{\fill}}c||ccc|ccc|ccc|ccc}
		\hline
        \multirow{2}{1.0cm}{\textbf{Variant}}&
		\multicolumn{3}{c|}{\textbf{PeMS}} &
        \multicolumn{3}{c|}{\textbf{Beijing}} &
        \multicolumn{3}{c|}{\textbf{Electricity}} &
		\multicolumn{3}{c}{\textbf{COVID-CHI}}
		\\
		\cline{2-13}
        \multicolumn{1}{c||}{} & 
		\multicolumn{1}{c}{\textit{RMSE}} & 
		\multicolumn{1}{c}{\textit{MAE}} &
		\multicolumn{1}{c|}{\textit{MAPE}} & 
        \multicolumn{1}{c}{\textit{RMSE}} & 
		\multicolumn{1}{c}{\textit{MAE}} &
		\multicolumn{1}{c|}{\textit{MAPE}} & 
        \multicolumn{1}{c}{\textit{RMSE}} & 
		\multicolumn{1}{c}{\textit{MAE}} &
		\multicolumn{1}{c|}{\textit{MAPE}} & 
		\multicolumn{1}{c}{\textit{RMSE}} & 
		\multicolumn{1}{c}{\textit{MAE}} & 
		\multicolumn{1}{c}{\textit{MAPE}}
		\\
		\hline
        Backbone & 2.385 & 1.125 & 1.976\% & 7.754 & 3.564 & 11.916\% & 97.087 & 51.684 & 16.120\% & 16.210 & 6.857 & 80.038\% \\
        RM & 2.359 & 1.114 & 1.975\% & 7.747 & 3.607 & 12.422\% & 69.945 & 36.093 & 12.411\% & 17.578 & 7.362 & 84.249\% \\
        RM+PM & 2.319 & 1.060 & \textbf{1.856\%} & 8.748 & 4.048 & 14.627\% & 71.184 & 36.652 & 12.875\% & 14.470 & 6.296 & 74.447\% \\
        MemDA & \textbf{2.297} & \textbf{1.053} & 1.861\% & 6.720 & \textbf{3.192} & \textbf{9.913\%} & \textbf{67.413} & \textbf{34.814} & \textbf{12.186\%} & \textbf{14.003} & \textbf{6.115} & \textbf{72.088}\% \\
		\hline
	\end{tabular*}}
\end{table*}

\subsection{Settings}
We annotate the time of events causing drift on each dataset. Based on this we divide the data before the drift into the training set and the rest as the test set, and further selected 20\% of the training set as the validation set. The cut-off positions for training and testing are shown in Fig. \ref{fig:data}. As for the model, the predicted leading time $\alpha$ is set to 12. The hyperparameters are chosen based on the performance of the validation set. The look-back days $K$ for drift embedding is set to 2, with the encoding dimension $C_e$ being 256, the number of prototypes $L$ of Pattern Memory being 20, and the dimension $D$ being 32. The GW-Net \cite{wu2019graph} is chosen as the encoder backbone for the model as its proven superior performance \cite{jiang2021dl}. We set the $N_s$ in the drift adaptation module as 5, and the meta-layer is implemented by a linear layer. As for the training, \textit{MAE} is chosen to be optimized using Adam, with a batch size set to 64 and a learning rate of 0.001. The best model is selected by early stop with a maximum of 200 epochs. The experiments were performed on a GPU server with four \textit{GeForce RTX 3090} graphic cards. Root Mean Square Error (\textit{RMSE}), Mean Absolute Error (\textit{MAE}), and Mean Absolute Percentage Error (\textit{MAPE}) are selected to evaluate the results.

\subsection{Baselines}
\subsubsection{Non-adaptive Models}
We implemented state-of-the-art methods on urban data prediction for comparison, including:
\begin{itemize}
    \item \textbf{CopyLastDay (CLD)}: Observations from the same moment of the previous day are used as predicted values.
    \item \textbf{ARIMA}: AutoRegressive Integrated Moving Average is a widely used statistical model for time series prediction by identifying patterns in data.
    \item \textbf{AGCRN} \cite{bai2020adaptive}: A node parameter learning and graph generation model for traffic forecasting tasks. It automatically captures the spatial and temporal correlation of covariates in time series data by adaptive modules.
    \item \textbf{DCRNN} \cite{li2018diffusion}: A model for modeling multivariate time-series prediction tasks using bidimensional graph diffusion and recurrent neural networks.
    \item \textbf{MTGNN} \cite{wu2020connecting}: An advanced multivariate time-series prediction model using active graph learning and multi-kernel TCN modules.
    \item \textbf{STGCN} \cite{yu2018spatio}: A spatio-temporal network that combines graph convolution with 1D convolution.
    \item \textbf{ASTGCN} \cite{guo2019attention}: A novel attention-based spatial-temporal graph convolution model. It features three temporally separate components to model periodic dependencies.
    \item \textbf{GMAN} \cite{zheng2020gman}: A model using multi-attentive networks with gated fusion to fit nonlinear temporal correlations and adaptively fuse information.
    \item \textbf{StemGNN} \cite{cao2020spectral}: A predictive model for modeling the spatial and temporal dependence of multivariate time series in the spectral domain.
    \item \textbf{GW-Net} \cite{wu2019graph}: A state-of-the-art multivariate modeling model using parametric graph inputs and a WaveNet-like temporal dilated structure.
    \item \textbf{STTN} \cite{xu2020spatial}: A model in which various relationships related to different factors in covariates are modeled jointly by spatial and temporal transformers.
    \item \textbf{EAST-Net} \cite{wang2022event}: A novel model for modeling event-induced non-stationary prediction problems in multimodal data via heterogeneous mobility information networks.
\end{itemize}

\subsubsection{Adaptive Models}
The latest advanced adaptive models for non-stationary time series are also chosen to compare the effect on the concept drift problem, including:
\begin{itemize}
    \item \textbf{RevIN} \cite{kim2021reversible}: An instance normalization method for against distribution shift that suppresses non-stationary information in one input layer and restores it in the output layer.
    \item \textbf{Dish-TS} \cite{fan2023dish}: An advanced version of RevIN, which designs a coefficient net that maps input sequences into learnable distribution to relieve distribution change.
\end{itemize}
For a fair comparison, we select adaptive models that can also function pluggable for comparison. The Backbone model is chosen as GW-Net based on its optimal performance on the four datasets.

\subsection{Performance Evaluation}
\subsubsection{Overall Performance Comparison}
Table \ref{tab:mse} shows the results of MemDA and baselines under three evaluation metrics. The proposed MemDA consistently outperforms the best baseline model on all four datasets, achieving a reduction in \textit{MAE} of approximately 6\% to 12\%. These results demonstrate the effectiveness of the adaptive model under drift scenarios. Furthermore, we observed that the model performance on test data varied. Among the two speed datasets, the Beijing dataset presented a greater challenge due to the impact of lifting lockdowns, which caused traffic speeds to drift from uniformly smooth to fluctuating. The proposed model exhibited a greater advantage over the baselines when dealing with such new patterns, resulting in an average improvement of 13\% for the three evaluation metrics. The PeMS dataset, on the other hand, drifted from congested to uncongested, which can be interpreted as an oversampling from the original distribution. Therefore, the concept drift problem may not cause significant issues, but our approach still achieved a 6\% improvement in the \textit{MAE} metric. For the Electricity datasets, ASTGCN, which also utilized periodically constructed inputs, achieved the best results among the baseline models. However, our proposed framework still achieves an additional 11\% improvement, demonstrating the potential of self-configuration capabilities in fully exploiting long-term historical information. Finally, all baseline models showed poor results on the COVID-CHI dataset, failing to match the simplest rule-based model CopyLastDay in terms of \textit{MAE}. Nevertheless, MemDA still stands out and shows strong robustness in the drift.

Table \ref{tab:adap_mse} shows the results of MemDA and SOTA adaptive models. The experimental results presented clearly demonstrate that our proposed MemDA model outperforms the other existing adaptive models. Moving onto the analysis of these results, RevIN and Dish-TS dynamically normalize the input data during neural network computation, which can only alleviate issues caused by changes in the marginal probability distribution $p(x)$. However, they overlook changes in the conditional probability distribution, which makes them disadvantaged in comparison. In addition, the Beijing and PeMS datasets exhibit significant variations in data scale (refer to Fig. \ref{fig:data}). Both RevIN and Dish-TS fail to show improvement in these scenarios, indicating a potential inability to handle drastic distribution drift effectively. In contrast, MemDA shows improved performance, demonstrating its robust prediction logic that can be particularly effective for prediction in challenging environments characterized by substantial concept drift.

\begin{figure}[h]
	\centering
	\includegraphics[width=0.47\textwidth]{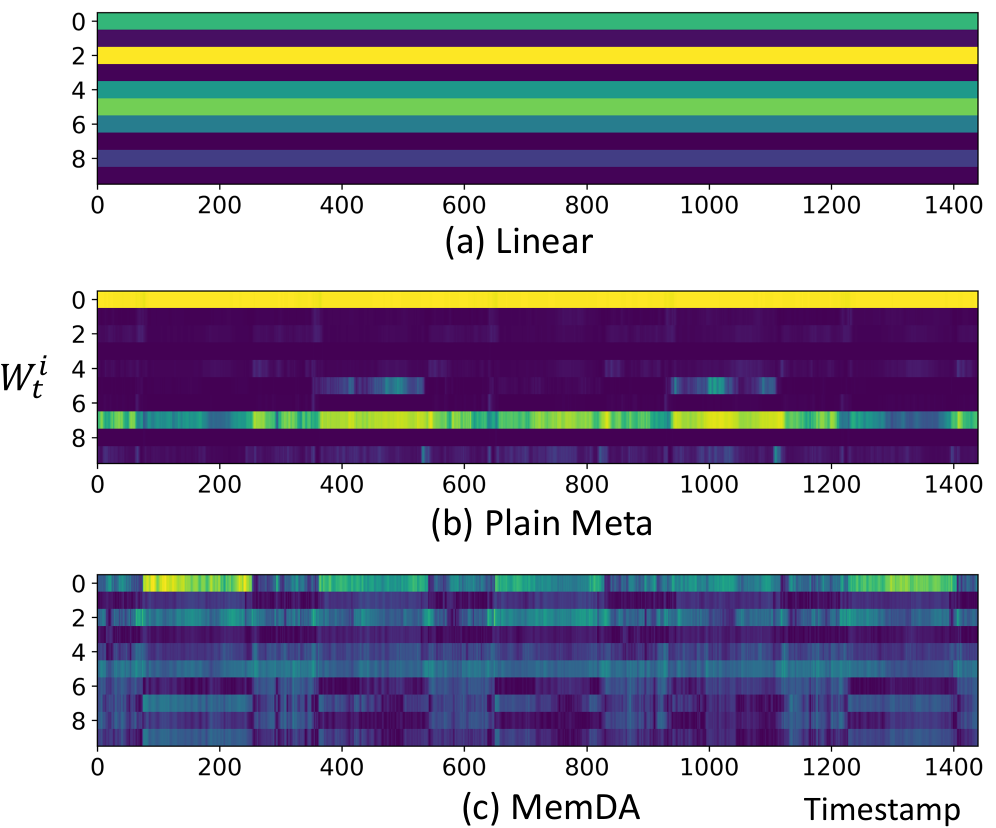}
	\caption{Visualization of the variation of the adaptation parameter $W_t$ on PeMS dataset. Where (a) is from the RM+PM, (b) is from the Meta, and (c) is from the proposed model.}
	\label{fig:adaptive}
\end{figure}

\subsubsection{Ablation Test}
A series of variant experiments are conducted to verify the effectiveness of each module. The variant with only the original encoder network (GW-Net) is referred to as the Backbone, to which we add RM and PM modules. The Meta variant removes the meta-dynamic layer and directly generates $W_t$ through a linear network connected to $E_t$. Results are shown in Table \ref{tab:ablation}, where we can observe that (1) Simply augmenting the long-term input via memory modules does not always improve performance, with negative effects for Beijing and COVID-CHI datasets. (2) Simply generating $W_t$ through the embedding of input, as is done in the Meta variant, does not lead to satisfactory results. In contrast, the MemDA evaluates drift by embedding pairs and generates $W_t$ accordingly, demonstrating a proven adaptation capability and the effectiveness of the meta-dynamic layer.

To interpret how MemDA copes with drift, we visualize the fusion parameter $W_t$ in Fig.~\ref{fig:adaptive}. Specifically, we show the solidified $W$ (the variant of RM+PM) in (a), $W_t$ generated by the Simple Meta module (the variant of RM+PM+Meta) in (b), and $W_t$ of MemDA in (c). The visualization results demonstrate that the MemDA module enables the adaptation model to be aware of changes in data distribution and to make proactive adjustments to the model parameters in response to different concepts at different times. In contrast, the solidified model can only maintain the prediction logic that fits the training set, and the Simple Meta variant, although making some adjustments in response to different concepts, demonstrates a weak adaptation capability that cannot effectively track drift.

\begin{figure}[h]
	\centering
	\includegraphics[width=0.45\textwidth]{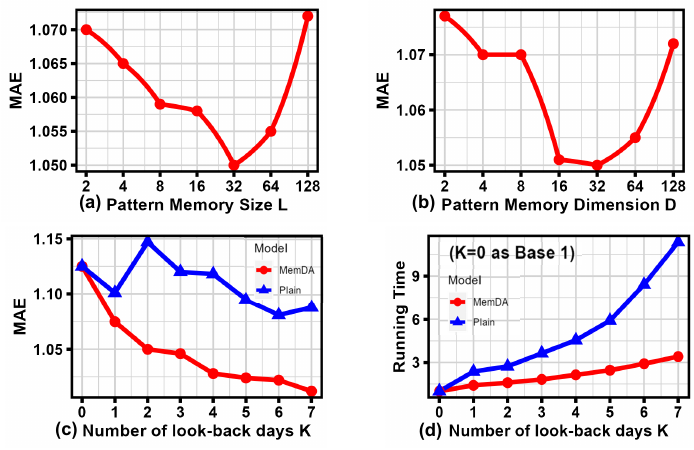}
	\caption{Sensitivity Analysis}
	\label{fig:sensitivity}
\end{figure}

\begin{figure*}[h]
	\centering
	\includegraphics[width=1.0\textwidth]{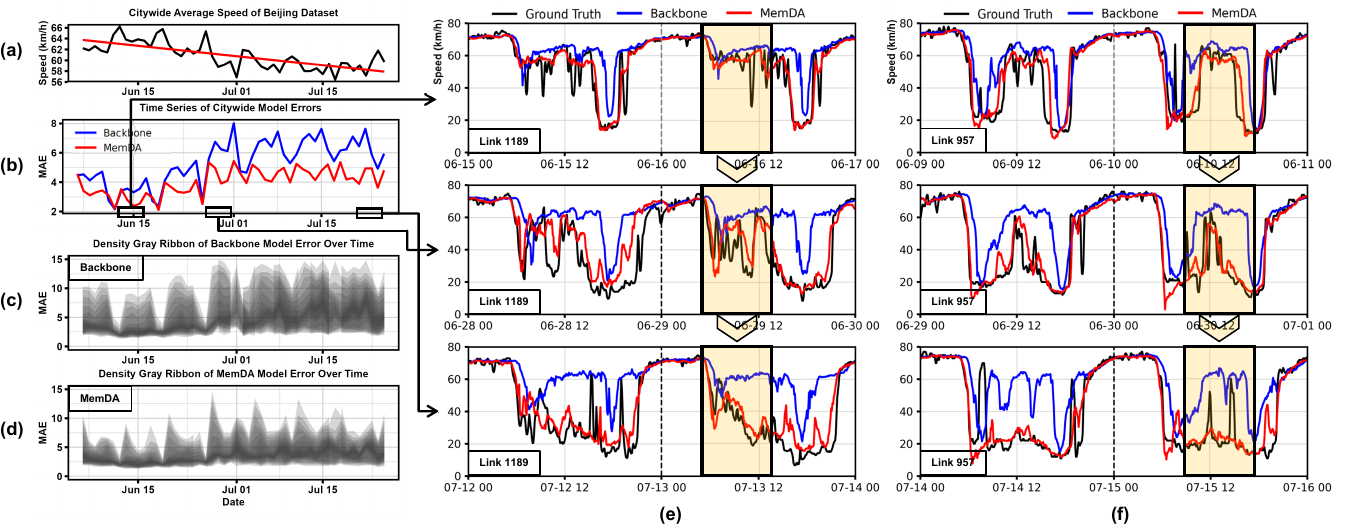}
	\caption{Case Study. (a) Average speed change of Beijing test set. (b) The daily error of Backbone and MemDA model. (c,d) Density error gray ribbon of Backbone/MemDA. (e,f) The model prediction results of two roads at different time periods in the test set.}
	\label{fig:case}
\end{figure*}

\subsubsection{Sensitivity Analysis}
We experimented the effects of Pattern Memory size $L$ and dimension $D$ of MemDA on the PeMS dataset. The results are presented in Fig. \ref{fig:sensitivity}(a,b). The model achieves optimal performance with a memory size of 32. This optimal point represents a balance between storing enough prototype patterns to accurately detect out-of-distribution instances and the ability to generalize by discarding noise. Similarly, optimal performance can also only be achieved with a suitable memory dimension.

We further tested the effect of look-back days $K$. As a comparison, we constructed a Plain model, in which naive uses multiple branches with a linear fusion layer to process the input segments from different days (same as Fig. \ref{fig:input_structure} (a)). The results are shown in Fig. \ref{fig:sensitivity}(c). MemDA demonstrates a continuous improvement of the effect with increasing look-back days, while the Plain model fluctuates and does not show significant trends. This demonstrates that dealing with concept drift is not a simple matter of adding inputs. And even sometimes the increase in look-back days isn't beneficial (Plain model 1 to 2). As the distribution of data changes over time, a model that heavily relies on past data could be misled by outdated information that no longer applies to the current. This is a risk when increasing input without an adaptive mechanism.

A comparison of the running speed of the Plain and MemDA is shown in Fig. \ref{fig:sensitivity}(d). It demonstrates that the computational pressure of the Plain model shows an exponential increase as the number of backtracking days increases, and it takes more than 10 times more time to look back at a week. While MemDA enhanced by Replay Memory presents an efficient ability of slow linear increase in computation time, which is crucial for enhancing model effects in many real-time scenarios.

\subsection{Generalizability Evaluation}
The properties of MemDA enable it to be easily integrated with existing urban data forecasting models. We utilize the backbone of three widely used models, MTGNN, GMAN, and GW-Net as the encoder to test the encoder-agnostic performance of the proposed model. The results are presented in Table \ref{tab:encoder}, which demonstrates the generalizability and applicability of MemDA across various models. 

\begin{table}[h]
	\centering
	\caption{Encoder-agnostic Performance Evaluation}
	\label{tab:encoder}
	\addtolength{\tabcolsep}{-0.8pt}
    \resizebox{.48\textwidth}{!}{
	\begin{tabular*}{10.8cm}{@{\extracolsep{\fill}}l|ccc|ccc}
		\hline
        \multirow{2}{*}{\textbf{Backbone}}&
		\multicolumn{3}{c|}{\textbf{Beijing}} &
        \multicolumn{3}{c}{\textbf{Electricity}}
		\\
		\cline{2-7}
        \multicolumn{1}{c|}{} & 
		\multicolumn{1}{c}{\textit{RMSE}} & 
		\multicolumn{1}{c}{\textit{MAE}} &
		\multicolumn{1}{c|}{\textit{MAPE}} & 
        \multicolumn{1}{c}{\textit{RMSE}} & 
		\multicolumn{1}{c}{\textit{MAE}} &
		\multicolumn{1}{c}{\textit{MAPE}}
		\\
		\hline
        MTGNN & 10.976 & 5.120 & 19.928\% & 96.475 & 51.562 & 15.410\% \\
        +MemDA & 9.411 & 4.379 & 16.923\% & 69.381 & 35.416 & 11.517\% \\
        $\Delta$\% & 14.26\% & 14.48\% & 15.08\% & 28.08\% & 31.31\% & 25.26\% \\
        \hline
        GMAN & 11.555 & 6.448 & 21.941\% & 138.436 & 81.285 & 26.263\% \\
        +MemDA& 10.005 & 6.011 & 19.041\% & 91.030 & 49.400 & 15.072\% \\
        $\Delta$\% & 13.41\% & 6.78\% & 13.22\% & 34.24\% & 39.23\% & 42.61\% \\
        \hline
        GW-Net & 7.754 & 3.564 & 11.916\% & 97.087 & 51.684 & 16.120\% \\
        +MemDA & 6.720 & 3.192 & 9.913\% & 67.413 & 34.814 & 12.186\% \\
        $\Delta$\% & 13.37\% & 10.43\% & 16.81\% & 30.56\% & 32.64\% & 24.40\% \\
		\hline
        	\end{tabular*}}
\end{table}

\section{Case Study}

We do case studies from both macro and micro perspectives. Fig. \ref{fig:case} (a,b,c,d) shows the city-wide average speed of the Beijing test dataset, the daily error time series of the Backbone (GW-Net) and MemDA model, the daily error density gray ribbon distribution of the Backbone and MemDA model, respectively. As the average road speed in Beijing dwindles, traffic conditions progressively worsen. This downturn significantly disrupts the predictability of traffic patterns, triggering a substantial decline in the performance of the Backbone model. This is reflected in the model's increasing mean prediction error, with a considerable spread ranging from low to high errors. In stark contrast, the incorporation of MemDA into the Backbone manages to keep the error distribution tightly controlled and centered on the lower values.

A microscopic analysis of the performance was focused on two representative sample roads selected from the initial, intermediate, and final phases of the test set. The zoomed-in views are shown in Fig. \ref{fig:case}(e,f). As can be seen, both roads with different speed patterns undergo significant pattern shifts over time (marked with yellow boxes), and the Backbone model just barely works near the training periods. However, as the concept drifts, Backbone's predictions stray further from the actual values. In essence, the Backbone model appears to be memorizing speed patterns rather than predicting them. The introduction of MemDA to it, however, imparts a remarkable transformation to the Backbone, enabling it to track the changes in data patterns and always maintain robustness after drift.

\section{Conclusion}
In this study, we target the problem of concept drift in urban time series data prediction. A general adaptive urban model leverages a dual-memory module, and a meta-dynamic network is designed to enable self-configuration. Extensive experiments demonstrate the effectiveness of the proposed framework. In the future, we aim to explore online approaches for constructing robust models as well as additional strategies for addressing spatial drift.

\begin{acks}
This work was supported by Japan Science and Technology Agency (JST) SPRING, Grant Number JPMJSP2108, and the 2022 Tencent Rhino-Bird Research Elite Program.
\end{acks}

\bibliographystyle{ACM-Reference-Format}
\bibliography{bibliography}


\end{document}